\documentclass[conference]{IEEEtran}

\usepackage{amsmath,amsfonts}
\usepackage{booktabs,mathtools,multicol,multirow,diagbox} 
\usepackage{array}
\usepackage[caption=false,font=normalsize,labelfont=sf,textfont=sf]{subfig}
\usepackage{textcomp}
\usepackage{stfloats}
\usepackage{url}
\usepackage{graphicx}
\usepackage{cite}

\newcommand{\paren}[1]{\left({#1}\right)}
\newcommand{\squares}[1]{\left[{#1}\right]}
\newcommand{\mb}[1]{\mathbf{#1}}
\DeclarePairedDelimiter{\norm}{\lVert}{\rVert}

\begin{document}

\title{Is Geometry Enough? An Evaluation of Landmark-Based Gaze Estimation}

\author{Daniele Agostinelli$^1$, Thomas Agostinelli$^1$, Andrea Generosi$^2$, Maura Mengoni$^1$ \\ \footnotesize
$^1$Department of Industrial Engineering and Mathematical Sciences, Università Politecnica delle Marche, via Brecce Bianche, 12, Ancona, 60131, Italy\\
$^2$Department of Science and Information Technology, Università Pegaso, piazza Trieste e Trento, 48, Napoli, 80132, Italy}

\maketitle

\begin{abstract}
Appearance-based gaze estimation frequently relies on deep Convolutional Neural Networks (CNNs). These models are accurate, but computationally expensive and act as ``black boxes", offering little interpretability. Geometric methods based on facial landmarks are a lightweight alternative, but their performance limits and generalization capabilities remain underexplored in modern benchmarks. In this study, we conduct a comprehensive evaluation of landmark-based gaze estimation. We introduce a standardized pipeline to extract and normalize landmarks from three large-scale datasets (Gaze360, ETH-XGaze, and GazeGene) and train lightweight regression models, specifically Extreme Gradient Boosted trees and two neural architectures: a holistic Multi-Layer Perceptron (MLP) and a siamese MLP designed to capture binocular geometry. We find that landmark-based models exhibit lower performance in within-domain evaluation, likely due to noise introduced into the datasets by the landmark detector. Nevertheless, in cross-domain evaluation, the proposed MLP architectures show generalization capabilities comparable to those of ResNet18 baselines. These findings suggest that sparse geometric features encode sufficient information for robust gaze estimation, paving the way for efficient, interpretable, and privacy-friendly edge applications. The source code and generated landmark-based datasets are available at: \url{https://github.com/daniele-agostinelli/LandmarkGaze.git}.
\end{abstract}

\begin{IEEEkeywords}
Gaze estimation, Facial landmarks, Machine learning, Deep learning, Pattern recognition
\end{IEEEkeywords}

\section{Introduction}
Gaze estimation is a critical task for applications in human-computer interaction, automotive safety, and healthcare monitoring. Common techniques are categorized mainly into model-based and appearance-based methods~\cite{Cheng2024}. Model-based methods rely on anatomical priors to reconstruct 3D eyeball structures, typically employing dedicated hardware such as infrared cameras and light sources~\cite{Guestrin2006,Hansen2009,Takemura2017}. Although these approaches achieve high accuracy, hardware requirements and high costs often limit their deployment. In contrast, appearance-based methods estimate gaze directly from images of the eyes or face, making them more suitable for unconstrained settings~\cite{Cheng2024}.

Since the introduction of deep learning for gaze estimation~\cite{Zhang2015}, the field moved from extractors of eye features (e.g., using asymmetric networks~\cite{Cheng2020}, pictorial features \cite{Park2018b}, or few-shot learning strategies~\cite{Park2019}) to models leveraging full-face context (e.g., spatial attention~\cite{Zhang2017}, dilated convolutions~\cite{Chen2019}, coarse-to-fine architectures \cite{Cheng2020b}, adaptive fusion \cite{Bao2021}, and transformers~\cite{Cheng2022}). More recent studies explored unsupervised domain adaptation \cite{Liu2021,Wang2022,Bao2022,Cai2023} and domain generalization \cite{Bao2024,Cheng2022b,Xu2023,Liang2024}. 

The main challenges in these approaches remain the high computational costs and lack of interpretability, since most of the appearance-based models rely on computationally heavy Convolutional Neural Network (CNN) backbones. 
These limitations might be overcome by adopting geometric models that estimate gaze directly from facial landmarks. While landmarks are typically used for initial data normalization~\cite{Zhang2018} or as auxiliary features that are combined with image data~\cite{Krafka2016, Yu2018, Lei2023}, some studies have explored the strategy of regressing gaze exclusively from landmarks. In contexts such as physical ergonomics or emotion recognition, similar approaches have succeeded to train efficient predictors on geometric features rather than raw image data, supporting the fact that structural information alone is sufficient for regression tasks (e.g.,~\cite{agostinelli2026, macedo2024, kumar2025}).

Early work in this direction suggested that 2D facial landmarks can be linked to 3D gaze through explicit geometric modeling, using 3D eye-face models~\cite{Chen2008,Wang2017}. Building on this idea, Park et al.~\cite{Park2018} detected detailed eye-region landmarks with a stacked-hourglass network and fed these features into a support vector regressor, explicitly combining learned landmark detection with traditional geometric estimation. Similarly, other studies proposed approaches to estimate gaze solely from landmarks in the eye region or relative facial keypoint coordinates extracted with pose estimators~\cite{Oh2022, Her2023, ye2023}. 

Despite these contributions, landmark-based gaze estimation remains underexplored and lacks a systematic assessment against modern appearance-based baselines. Most prior studies train on different synthetic data (e.g., UnityEyes~\cite{Wood2016}) and validate their methods in specific settings. As a result, it remains unclear to what extent landmarks alone can be used for accurate and robust gaze estimation, and what are the capabilities and the current limitations of this approach. 

In this study, we propose a pipeline to derive normalized landmark-based datasets from annotated face images available in the literature, i.e., Gaze360~\cite{Kellnhofer2019}, ETH-XGaze~\cite{Zhang2020}, and GazeGene~\cite{Bao2025}. After analyzing the coverage of these datasets, we use them to train and evaluate Extreme Gradient Boosted decision trees (XGBoost) and Multi-Layer Perceptrons (MLPs), which we interpret by Permutation Feature Importance (PFI). We find that lightweight MLPs can achieve cross-domain performances comparable to image-based baselines, suggesting that facial landmarks contain the information necessary for accurate and robust gaze estimation. In conclusion, we identify the precision of the landmark detector and the quality of the datasets as the main bottlenecks of this approach.

\section{Methods}
In this study, we propose and assess a framework for gaze estimation based only on landmarks. 
In Section~\ref{sec:dataset extraction} we develop a procedure to extract landmark-based datasets from three suitable datasets that are available in the literature: Gaze360~\cite{Kellnhofer2019}, ETH-Xgaze~\cite{Zhang2020}, and GazeGene~\cite{Bao2025}. 
In Section~\ref{sec:architecture and training}, we describe the proposed network architectures, training procedures and evaluation protocols.

\subsection{Extraction of landmark-based datasets}\label{sec:dataset extraction}
\begin{figure*}[!ht]
    \centering
    \includegraphics[width=5in]{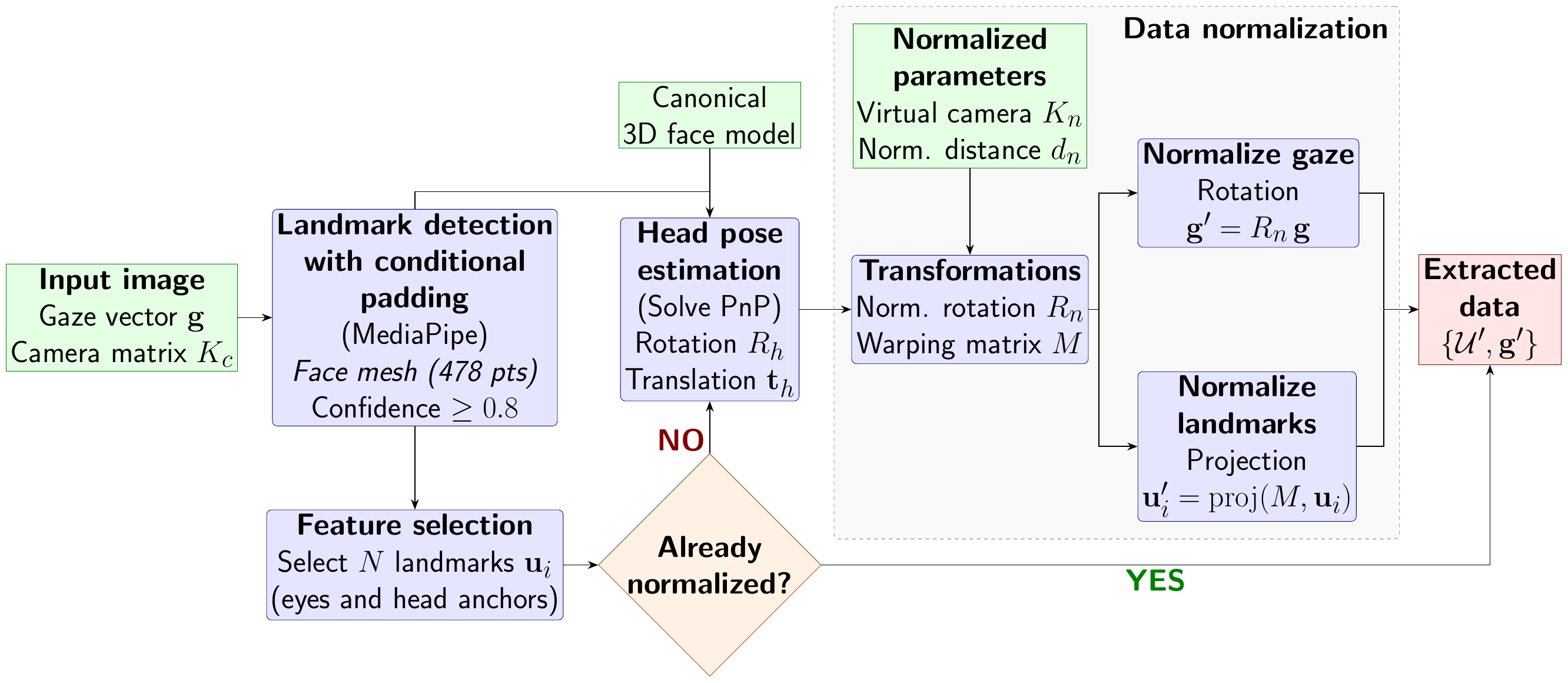}
    \caption{Pipeline for the extraction of landmark-based dataset. Images are processed to extract landmarks and head pose, which are then used to normalize the data into a virtual camera space.}
    \label{fig:dataset_pipeline}
\end{figure*}

We extract normalized landmark-based datasets from labeled images following the pipeline summarized in Fig.~\ref{fig:dataset_pipeline}. For each image, we detect dense facial landmarks, estimate the head pose, and normalize the features together with ground-truth gaze labels.

\subsubsection{Facial landmark detection}\label{sec:landmark detection}
\begin{figure}[h!]
    \centering
    \subfloat[]{\includegraphics[width=1.25in]{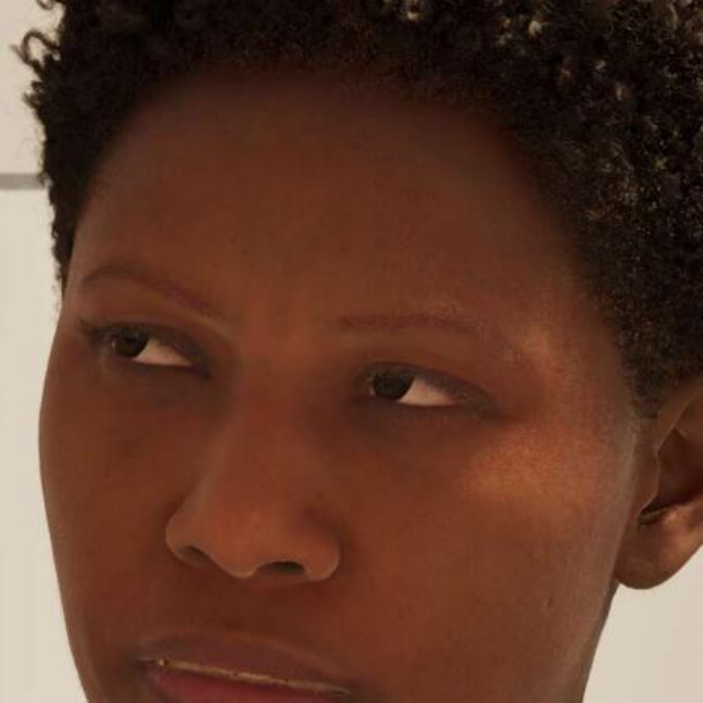}\label{fig:20_landmarks_a}}%
    \hfil
    \subfloat[]{\includegraphics[width=1.25in]{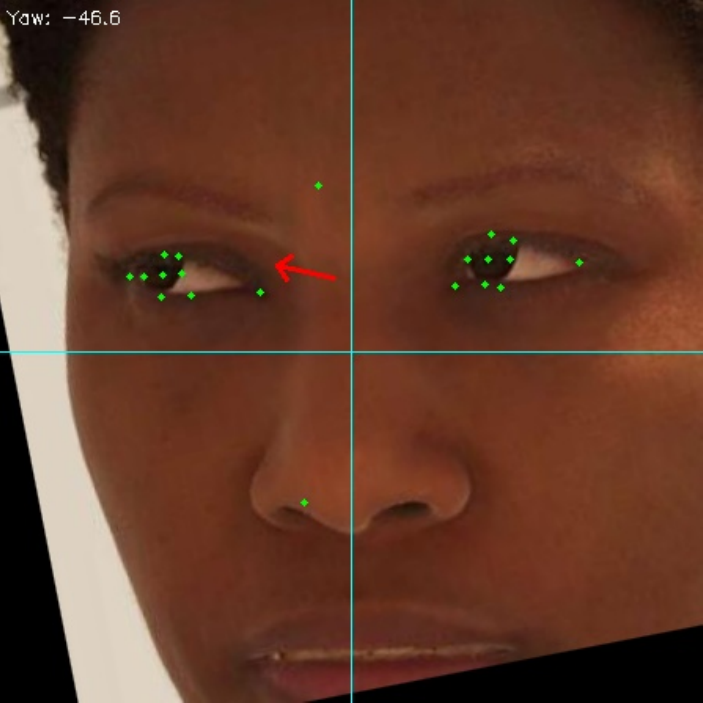}\label{fig:20_landmarks_b}}%
    \caption{Example image from the GazeGene dataset~\cite{Bao2025}: (a) original and (b) normalized images. In (b), the red arrow indicates the normalized gaze direction, cyan lines are the image principal axis, and green circles denote the $N=20$ landmarks extracted from the MediaPipe face mesh~\cite{Lugaresi2019} to represent gaze and head orientation: the right iris (indices 473--477) and eye contour (263, 362, 374, 386), the left iris (468--472) and eye contour (33, 133, 145, 159), and two head anchors given by the nose tip (1) and the glabella (9).}
    \label{fig:20_landmarks}
\end{figure}

We detect faces and extract landmarks using MediaPipe~\cite{Lugaresi2019}, which generates a dense mesh of 478 points, including 10 landmarks of pupils and irises. We discard detections with a confidence score below $0.8$. 
To reduce detection failures on cropped face images (such as those in the GazeGene~\cite{Bao2025} and ETH-Xgaze datasets~\cite{Zhang2020}), we apply symmetric black padding to the image (extending each side by 25\% of the original dimensions). 
From the dense face mesh, we select a subset of $N=20$ landmarks critical for gaze estimation: two stable head anchors (nose tip and glabella) and, for each eye, the pupil center, four iris extrema, and four eye contour landmarks (corners and eyelids extrema), as shown in Fig.~\ref{fig:20_landmarks}. We denote by $\mathcal{U} = \{\mathbf{u}_i\}_{i=1}^N$ the set of these 2D landmarks, where $\mathbf{u}_i = [u_{i,x}, u_{i,y}]^\top$ in the raw image coordinate system.

\subsubsection{Head pose estimation}\label{sec: head pose estimation}
In each image, we estimate the head pose by aligning the detected landmarks to a canonical 3D face model with origin at the midpoint between the eyes and nose corners~\cite{Marchand2016}. In particular, we use \texttt{OpenCV}~\cite{opencv_library} to solve the Perspective-n-Point (PnP) problem to determine the rotation matrix $R_h = \squares{\mb{e}_1^{(h)}\vert \mb{e}_2^{(h)}\vert \mb{e}_3^{(h)}}$ and the translation vector $\mathbf{t}_h$ that map the 3D face model to the camera coordinate system.

\subsubsection{Data normalization}\label{sec:data normalization}
To remove variability in face distance, position, and in-plane rotation, we map data to a normalized space following the procedure described in~\cite{Zhang2018}.
Using the estimated head pose $\paren{R_h, \mb{t}_h}$, we construct the normalized coordinate system 
\begin{equation}\label{eq:normalized rotation matrix}
\mb{e}_1^{(n)} = \mb{e}_2^{(n)} \times \mb{e}_3^{(n)}, \quad \mb{e}_2^{(n)} = \frac{\mb{e}_3^{(n)} \times \mb{e}_1^{(h)}}{\norm{\mb{e}_3^{(n)}\times \mb{e}_1^{(h)}}}, \quad \mb{e}_3^{(n)} = \frac{\mb{t}_h}{\norm{\mb{t}_h}}, 
\end{equation}
where $\norm{\cdot}$ denotes the Euclidean norm.
Then we can map the original image plane to the normalized image plane through the perspective warping matrix
\begin{equation}\label{eq:warping homography}
    M = K_n S R_n K_c^{-1},
\end{equation}
where $K_c$ is the intrinsic matrix of the physical camera (either provided in the dataset or estimated from the image size), $R_n = \squares{\mb{e}_1^{(n)}\vert \mb{e}_2^{(n)} \vert \mb{e}_3^{(n)}}^\top$ is the rotation matrix to the normalized coordinate system, $S = \operatorname{diag}\paren{\frac{\norm{\mb{t}_h}}{d_n}, \frac{\norm{\mb{t}_h}}{d_n},1}$ scales the face to a fixed distance $d_n$, and $K_n$ is the intrinsic matrix of the virtual camera defined by focal length $f_n$ and principal point $\paren{c_x,c_y}$. 
In conclusion, for each raw image, we determine the normalized landmarks, $\mathcal{U}'=\left\lbrace \mathbf{u}'_i \right\rbrace_{i=1}^{N}$, via perspective projection of the raw landmarks $\mathcal{U}$, i.e., 
\begin{equation}
    \mathbf{u}'_i = \left[ \frac{v_{i,x}}{v_{i,z}}, \frac{v_{i,y}}{v_{i,z}} \right]^\top, \qquad
    {\mathbf{v}_i} = M \begin{bmatrix} {u}_{i,x} \\ {u}_{i,y} \\ 1 \end{bmatrix}.
\end{equation}
Analogously, we rotate the 3D ground-truth gaze vector $\mathbf{g}$ into the normalized frame as
\begin{equation}\label{eq:normalized gaze vector}
    \mathbf{g}' = R_{n} \, \mathbf{g}.
\end{equation}

In this study, we configure the normalized camera with an image resolution of $448 \times 448$ pixels ($c_x=c_y=224$), a focal length $f_n= 960$, and a normalized distance $d_n=300$ mm (Fig.~\ref{fig:20_landmarks}).

\subsubsection{Dataset-specific processing}\label{sec:dataset-specific processing}

\paragraph{Gaze360~\cite{Kellnhofer2019}}
This dataset consists of raw images with variable resolution, for which ground-truth head poses are not provided. We first detect facial landmarks (without padding) and estimate the head pose by refining an initial Efficient PnP (EPnP) solution via the Levenberg-Marquardt method.
Then, we normalize landmarks and ground-truth gaze vectors as detailed in Section~\ref{sec:data normalization}; to this aim, we approximate the camera intrinsics by a pinhole camera model with the focal length set to the image width and the principal point at the image center.

\paragraph{ETH-XGaze~\cite{Zhang2020}}
We use the pre-processed version of face images with a resolution of $448 \times 448$ pixels, which are already normalized according to the procedure described in Section~\ref{sec:data normalization}. Consequently, we bypass the head pose estimation and normalization steps, detecting the normalized landmark directly on the images (with padding). Following~\cite{Zhang2020}, we apply histogram equalization to the Y-channel for frames with an index greater than $524$ to ensure consistency across varying lighting conditions.
Finally, we convert the ground-truth labels, originally provided as pitch and yaw, into 3D unit vectors in the normalized camera coordinate system. 

\paragraph{GazeGene~\cite{Bao2025}}
For this dataset, we implement the full processing pipeline: facial landmark detection (with padding), head pose estimation, and data normalization as described in Secs.~\ref{sec:landmark detection},~\ref{sec: head pose estimation} and~\ref{sec:data normalization}. We estimate the head pose by solving the Perspective-n-Point (PnP) problem using the \texttt{OpenCV} iterative solver. We initialize the solver with the ground-truth head pose provided by the dataset, but we compute the normalization matrix $R_n$ by the re-estimated pose to align the normalization warp and the detected features.

\subsection{Landmark-based gaze estimation}\label{sec:architecture and training}
In this study, we propose and evaluate three regression models to map the $N=20$ normalized landmarks, $\mathcal{U}'$, to the 3D gaze vector in the normalized camera coordinate system, $\mathbf{g}'  \in \mathbb{R}^3$: a holistic MLP, a siamese MLP, and an XGBoost decision tree. Figure~\ref{fig:architectures_diagram} shows the diagrams of the three architectures.

\subsubsection{Feature representation}
We transform the normalized coordinates $\mathcal{U}'$ into feature vectors, using geometric centering and scaling.

\paragraph{Global features ($\mathbf{f}_G$)}
For models processing the face as a whole, i.e., holistic MLP and XGBoost, we compute a single reference centroid $\mathbf{c}_{E}$ as the midpoint of the four eye corners. We center all landmarks relative to $\mathbf{c}_{E}$ and scale them by the normalized image width $w=448$, resulting in the feature vector $\mathbf{f}_{G} \in \mathbb{R}^{40}$ that preserves the relative geometry of the entire face.

\paragraph{Local features ($\mathbf{f}_L$, $\mathbf{f}_R$, $\mathbf{f}_H$, $\Delta \mathbf{c}$)}
For the siamese MLP, we define the centroids of the left and right eye corners, $\mathbf{c}_{L}$ and $\mathbf{c}_{R}$. For each eye there are $N_{e}=9$ landmarks, which are centered relative to their respective centroid and scaled by $w$, yielding two local feature vectors, $\mathbf{f}_{L}, \mathbf{f}_{R} \in \mathbb{R}^{18}$. The two remaining landmarks (head anchors) are concatenated in the vector $\mathbf{f}_H\in \mathbb{R}^4$. Finally, we compute the relative position vector $\Delta \mathbf{c}= \paren{\mathbf{c}_R-\mathbf{c}_L}/w$.

\subsubsection{Neural architectures}
We design two deep learning architectures sharing a common \textit{residual block} structure with hidden width $D$, which maps $\mathbf{h} \in \mathbb{R}^D$ into $\mathbf{y} = \mathbf{h} + \mathcal{F}(\mathbf{h}) \in \mathbb{R}^D$, where the residual function $\mathcal{F}(\cdot)$ is the sequence shown in Fig.~\ref{fig:architectures_diagram}d, namely, Linear($D{\to}D$) $\to$ BatchNorm $\to$ GELU $\to$ Dropout($p$) $\to$ Linear($D{\to}D$) $\to$ BatchNorm $\to$ GELU $\to$ Dropout($p$). We report the specific configurations for each network in Tab.~\ref{tab:model_configurations}.

\paragraph{Holistic MLP}
This network processes facial features through a monolithic approach. The architecture consists of an input projection layer from the vector $\mathbf{f}_{G}$ to a hidden width $D=256$ (Linear($2N{\to}D$) $\to$ BatchNorm $\to$ GELU), followed by a stack of $K=3$ residual blocks, and a regression head that reduces the dimension to $D/2$ before predicting the final 3D gaze vector (Linear($D{\to}D/2$) $\to$ GELU $\to$ Linear($D/2{\to}3$)). Figure~\ref{fig:architectures_diagram}a shows a diagram of this architecture.

\paragraph{Siamese MLP}
This architecture models the binocular nature of gaze using two parallel branches and a fusion stage (Fig.~\ref{fig:architectures_diagram}b). Two independent encoders process $\mathbf{f}_{L}$ and $\mathbf{f}_{R}$ in parallel, each mirroring the structure of the holistic model: an input projection layer followed by $K=3$ residual blocks with width $D=64$. To account for the spatial relationship between landmarks, we concatenate the latent vectors from the eye encoders with the relative position vector, $\Delta \mathbf{c} = \paren{\mathbf{c}_{R}-\mathbf{c}_{L}}/w \in \mathbb{R}^2$, and the normalized coordinates of the head anchors, $\mathbf{f}_H\in \mathbb{R}^4$. Finally, a fusion MLP predicts the gaze vector from these features (Fig.~\ref{fig:architectures_diagram}b).

\begin{table*}[!ht]
    \centering
    \caption{Summary of input representation and specific configurations of the MLPs.}
    \label{tab:model_configurations}
    \resizebox{5in}{!}{
    \begin{tabular}{l c c c}
        \toprule
        \textbf{Feature} & \textbf{Holistic MLP} & \textbf{Siamese MLP} \\
        \midrule
        \textbf{Input representation} & Global landmarks ($\mathbf{f}_{G}$) & Local eyes ($\mathbf{f}_{L}, \mathbf{f}_{R}$), geometry ($\Delta \mathbf{c}, \mathbf{f}_H$) \\
        \textbf{Hidden width (D)} & 256 & 64 (per branch) \\
        \textbf{N. residual blocks (K)} & 3  & 3 (per branch)  \\  
        \textbf{Dropout (p)} & 0.1 & 0.1\\
        \bottomrule
    \end{tabular}
    }
\end{table*}

\subsubsection{XGBoost}
\begin{figure*}[!ht]
    \centering
    \includegraphics[width=5in]{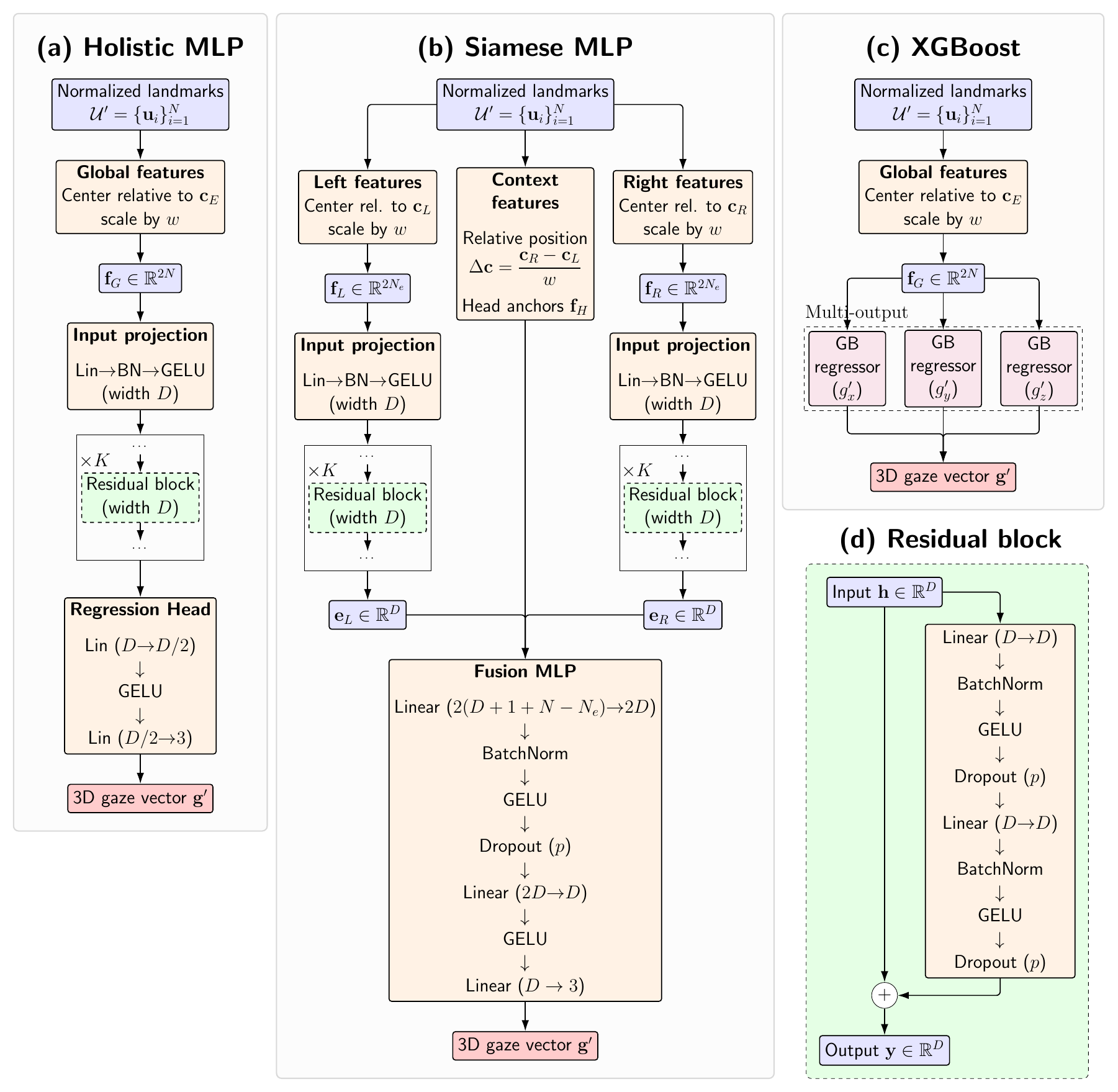}
    \caption{Architectural comparison of gaze estimation models. (a) Holistic Multi-Layer Perceptron (MLP): facial landmarks are processed through an input projection layer, a stack of $K$ residual blocks and a final regression head. (b) Siamese MLP: facial landmarks are split into local eye regions, processed by two independent encoders, and fused with geometric context (relative eye positions $\Delta\mathbf{c}$ and head anchors $\mathbf{f}_H$) via a fusion MLP. (c) XGBoost: a gradient-boosted tree approach using a multi-output regressor for the gaze vector components, $\mathbf{g}'= \paren{g'_x,g'_y,g'_z}$, based on global landmark features, $\mathbf{f}_G$. (d) Residual block used in (a) and (b), featuring a residual connection around two sets of Linear, BatchNorm, GELU, and Dropout layers.}
    \label{fig:architectures_diagram}
\end{figure*}

To establish a non-deep learning baseline, we train an extreme gradient boosted (XGBoost) regressor using the global features $\mathbf{f}_{G}$. Unlike monolithic neural networks, this architecture employs an ensemble of decision trees trained sequentially.
Since gradient boosting regressors typically predict scalar values, we employ a multi-output strategy, wrapping three independent estimators for the three components of the gaze vector, $\mathbf{g}'= \paren{g'_x, g'_y,g'_z}$, as shown in Fig.~\ref{fig:architectures_diagram}c. Each estimator is an additive model composed of $K=1000$ decision trees, where each tree has a maximum depth of $6$.

\subsubsection{Training and evaluation}
We train the MLPs using an \texttt{AdamW} optimizer (weight decay of $10^{-4}$, batch size of $64$) and a \texttt{ReduceLROnPlateau} scheduler (factor of $0.5$, patience of $4-5$ epochs) starting from a value of $10^{-1}$ for the holistic model and $10^{-3}$ for the siamese one. Both networks are trained for a maximum of $200$ epochs with early stopping (patience of 15 epochs). We define the loss function as the angular error between the predicted gaze vector ${\mathbf{g}}'_p$ and the ground-truth $\mathbf{g}'$, i.e., 
\begin{equation}\label{eq:loss function}
    E\paren{\mathbf{g}', {\mathbf{g}}'_p} = \arccos\paren{\frac{{\mathbf{g}}'_p \cdot \mathbf{g}'}{\norm{{\mathbf{g}}'_p} \norm{\mathbf{g}'}}}.
\end{equation}
This loss function maximizes cosine similarity, ensuring that the gaze direction is accurate regardless of the magnitude.\\

We train the XGBoost model using the histogram-based tree method, minimizing the mean squared error, with a learning rate of $0.05$ and a subsampling of 80\% of training instances and 80\% of features per tree.\\

To evaluate the generalization capability of the models, we split each dataset into training ($\sim$80\%), validation ($\sim$10\%), and testing ($\sim$10\%) sets. In particular, for GazeGene, we use subjects 1-46 for training, 47-51 for validation and 52-56 for testing; for ETH-Xgaze, we randomly split the 80 annotated subjects into 64 for training, 8 for validation, and 8 for testing; for Gaze360, we adopt the same splitting as in the original study~\cite{Kellnhofer2019}. First, we train models using training and validation sets. Then, we evaluate model performances in terms of the Mean Angular Error on the testing set of the same dataset used for training (within-domain evaluation), and on the entirety of the other datasets (cross-domain evaluation).

\section{Results}
\label{sec:results}
In this section, we present the quantitative results of the proposed landmark-based framework. In Section~\ref{sec:res_data}, we analyze the extracted landmark datasets compared to the original ones. In Section~\ref{sec:res_within}, we evaluate the within-domain accuracy of the proposed regression models (holistic and siamese MLPs, and XGBoost). In Section~\ref{sec:res_cross}, we present a cross-domain evaluation to assess generalization capabilities, comparing our lightweight models with the ResNet18 convolutional neural network. In Section~\ref{sec:feature importance}, we analyze the results of the training process using the Permutation Feature Importance method.

\subsection{Analysis of extracted datasets}
\label{sec:res_data} 
We report in Tab.~\ref{tab:retention} the retention rates of the face detection process using MediaPipe~\cite{Lugaresi2019} with a minimum confidence of $0.8$, as described in Section~\ref{sec:landmark detection}. For all datasets, more than 70\% of the images were processed successfully, with the highest rate achieved on GazeGene (82.61\%).

\begin{table}[!ht]
    \centering
    \caption{Dataset retention statistics. The retention rate indicates the percentage of images where MediaPipe successfully detected the facial landmarks with confidence $\ge 0.8$.}
    \label{tab:retention}
    \resizebox{.45\textwidth}{!}{
    \begin{tabular}{lccc}
    \toprule
    Dataset & Original samples & Extracted samples & Retention rate \\
    \midrule
    Gaze360~\cite{Kellnhofer2019} & $169935$ & $132651$ & $78.06$\%\\
    ETH-XGaze~\cite{Zhang2020} & $756540$ & $559057$ & $73.90$\%\\
    GazeGene~\cite{Bao2025} & $1008000$ & $832687$ & $82.61$\% \\
    \bottomrule
    \end{tabular}}
\end{table}

\begin{figure*}[!ht]
    \centering
    \subfloat[]{\includegraphics[width=3.5in]{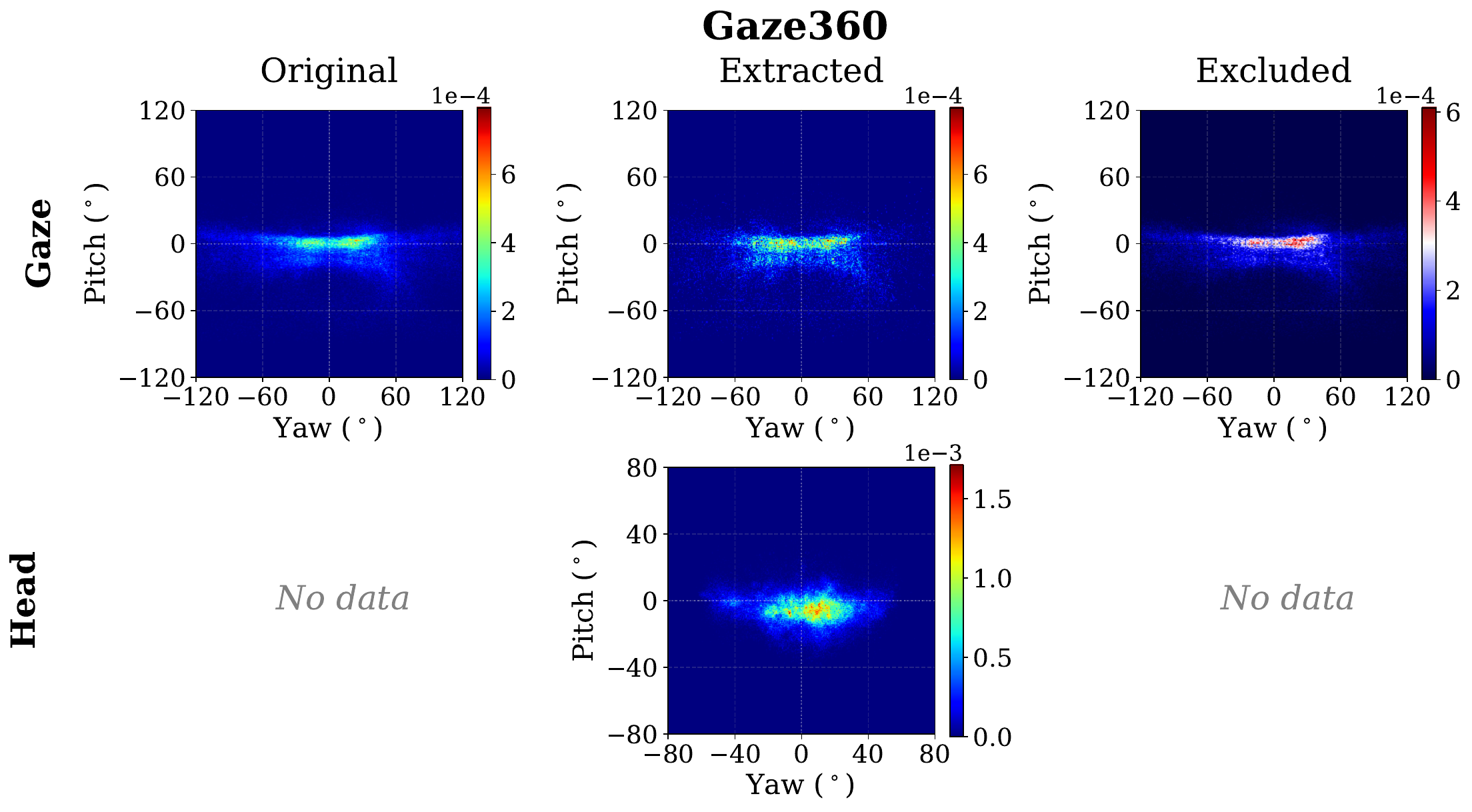}\label{fig:distr_gaze360}}
    \subfloat[]{\includegraphics[width=3.5in]{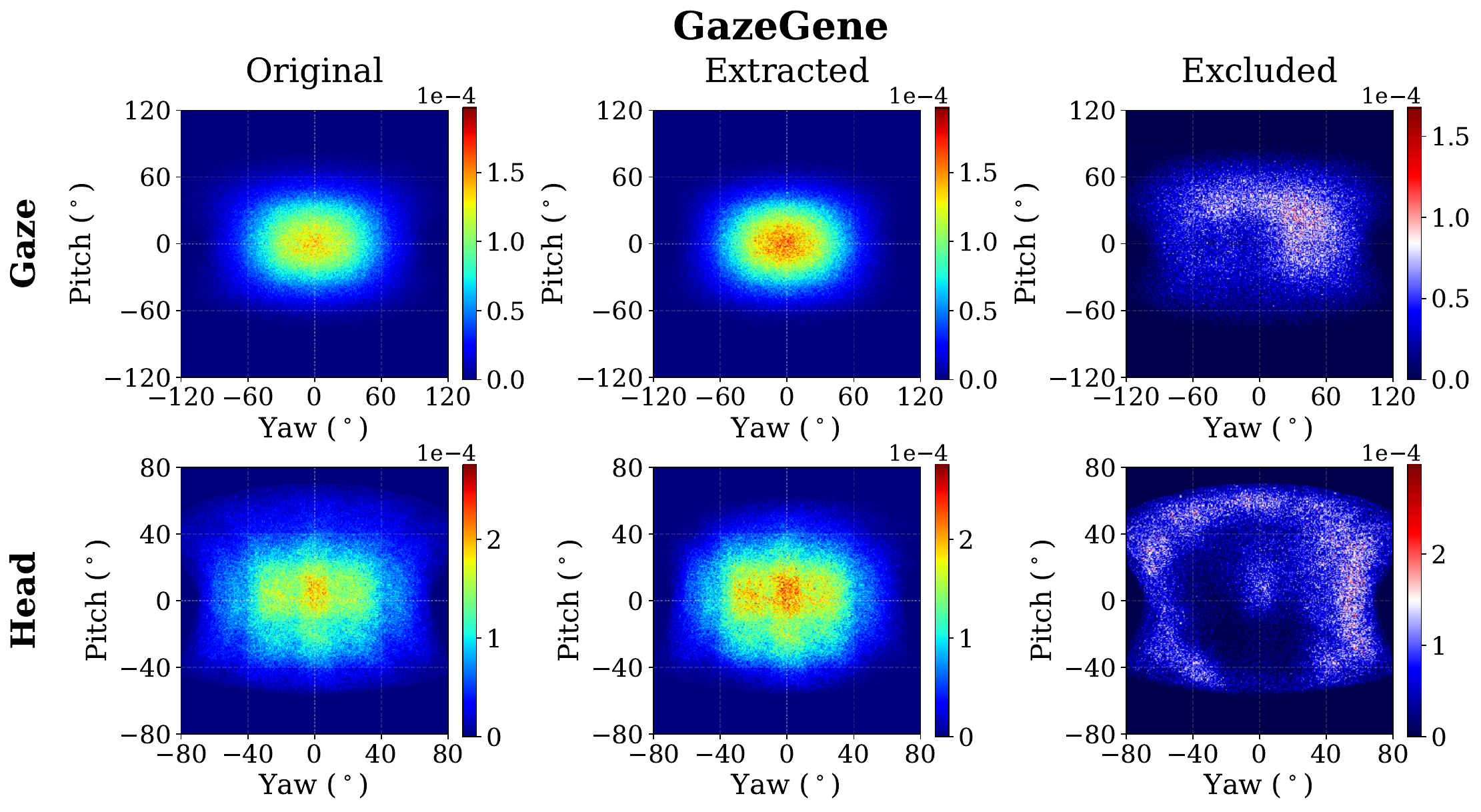}\label{fig:distr_gazegene}}\\
    \subfloat[]{\includegraphics[width=3.5in]{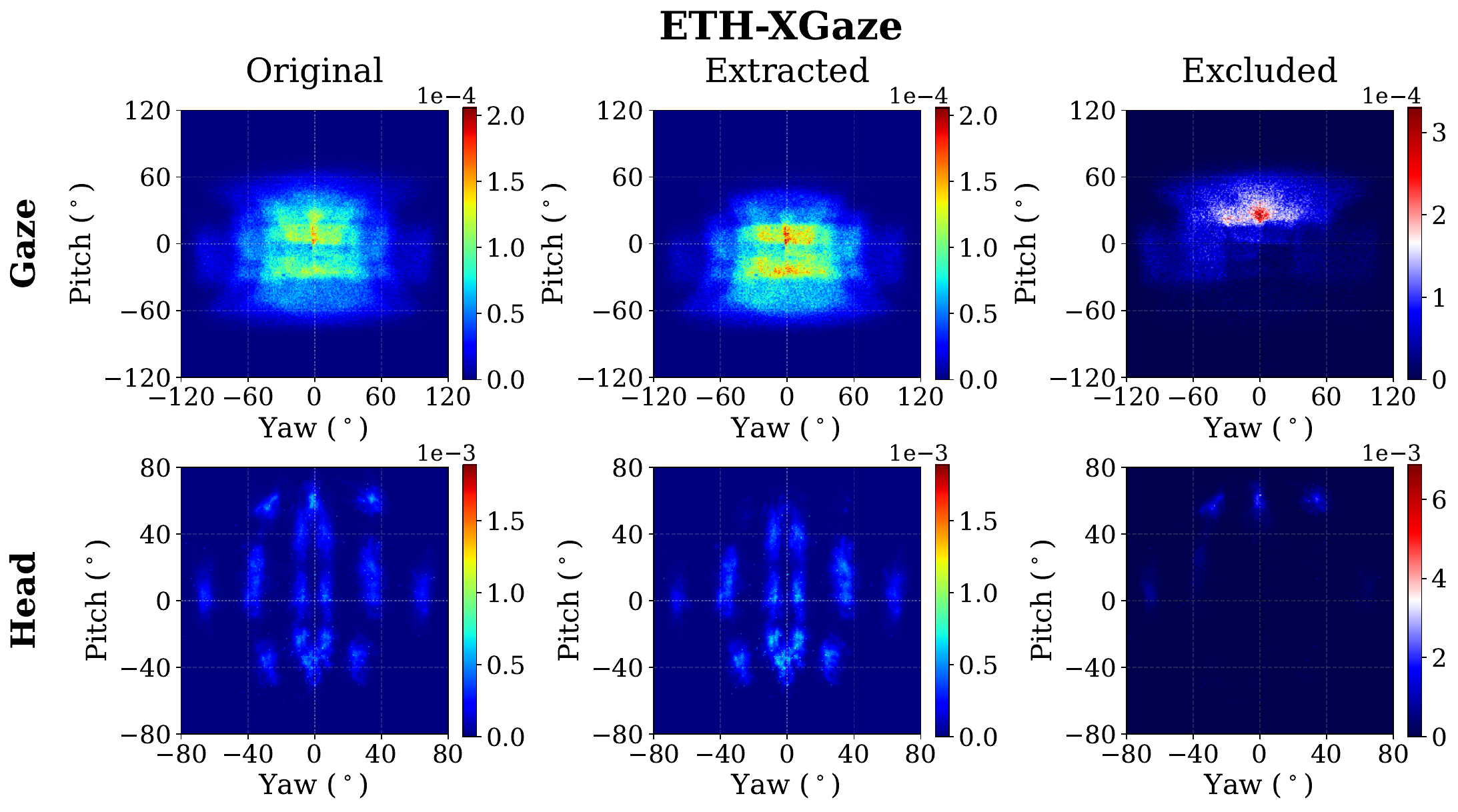}\label{fig:distr_ethxgaze}}
    \caption{Distributions of yaw and pitch angles (in degrees) for gaze (first row) and head pose (second row) across the three datasets - Gaze360 (a), GazeGene (b), and ETH-XGaze (c). Columns (left to right) show the distributions of the original, extracted, and excluded samples. Colors represent density values according to the reported colorbars.}
    \label{fig:distributions}
\end{figure*}

To investigate the geometric bias introduced by the extraction process, we compare the density distributions of gaze and head angles before and after processing. Figure~\ref{fig:distributions} illustrates such distributions for all datasets, together with the density of the excluded samples, showing that the landmark detector is robust for central gaze directions but tends to fail at extreme angles. This implies a reduction in the effective ranges of the extracted datasets compared to the original ones.

\subsection{Within-domain evaluation}
\label{sec:res_within}
\begin{figure}[!ht]
    \centering
    \includegraphics[width=2.5in]{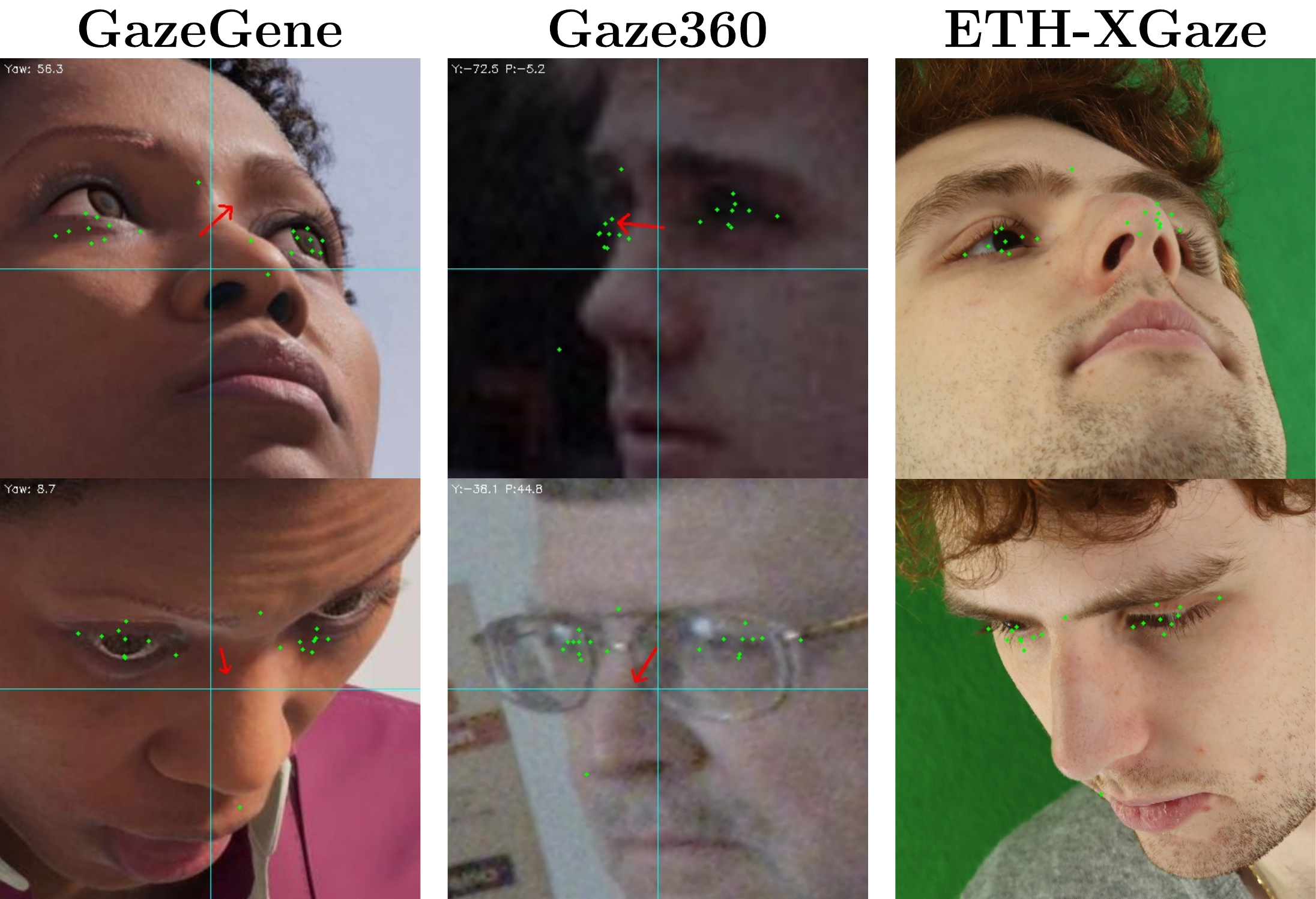}
    \caption{Examples of evident errors in face-landmark detection from different datasets (GazeGene~\cite{Bao2025}, Gaze360~\cite{Kellnhofer2019}, ETH-XGaze~\cite{Zhang2020}). Green circles denote the $N=20$ landmarks extracted using MediaPipe~\cite{Lugaresi2019}. These errors are likely due to low image resolution, face cropping, or extreme poses.}
    \label{fig:mediapipe_failures}
\end{figure}

In Tab.~\ref{tab:within_domain}, we compare the within-domain performance of the proposed models (holistic and siamese MLPs, and XGBoost) against the results reported in the literature for a ResNet18 architecture~\cite{Cheng2024,Bao2025}, representing a standard baseline for convolutional neural networks.
Both MLPs consistently performed better than the XGBoost model, with the siamese MLP achieving the best results among all the landmark-based methods. Notably, the ResNet18 baseline outperforms all other methods in the within-domain evaluation. This performance gap might be due to noise introduced in the extracted datasets, resulting from inaccuracies in the detection of facial landmarks and, consequently, head pose estimation. Indeed, qualitative inspection of the processed images reveals evident landmark detection errors (Fig.~\ref{fig:mediapipe_failures}), suggesting that the performances of these models are limited by the precision of the extracted datasets. 

\begin{table}[!ht]
    \centering
    \caption{Results of within-domain evaluation. Performance for landmark-based models is presented as Mean Angular Error and Standard Deviation (SD) in degrees ($^\circ$), calculated across all tested samples. Results for the ResNet18 are reported from the literature~\cite{Bao2025}. Bold values indicate the best performance.}
    \label{tab:within_domain}
    \resizebox{.45\textwidth}{!}{
    \begin{tabular}{l|c|ccc}
    \toprule
    & \textbf{Image-based} & \multicolumn{3}{c}{\textbf{Landmark-based}} \\
    \textbf{Dataset} & ResNet18 & XGBoost & Holistic MLP & Siamese MLP \\
    \midrule
    GazeGene  & \textbf{3.76} & 13.49 (8.71) & 9.00 (7.10) & 8.91 (7.02) \\
    ETH-XGaze & \textbf{5.73} & 11.63 (7.81) & 8.54 (6.71) & 8.50 (6.59) \\ 
    Gaze360   & \textbf{12.23} & 18.81 (14.38) & 14.95 (12.66) & 14.68 (12.70) \\ 
    \bottomrule
    \end{tabular}}
\end{table}

\subsection{Cross-domain evaluation}
\label{sec:res_cross}
We report in Tab.~\ref{tab:cross_domain} the results of the cross-domain evaluation.
Both landmark-based MLPs demonstrate generalization capabilities comparable to the ResNet18 baseline~\cite{Bao2025}, while the XGBoost architecture shows consistently lower performances. In fact, while ResNet18 generally achieves lower errors in within-domain settings, its performance degrades significantly in cross-domain settings.
In contrast, landmark-based models exhibit a relatively lower performance drop.
The siamese MLP maintains the highest accuracy among landmark methods in all cross-domain scenarios. 

\begin{table*}[!ht]
    \centering
    \caption{Results of cross-domain evaluation. For each model architecture, rows and columns represent the training and testing datasets, respectively. Performance for landmark-based models are presented as MAE (SD) in degrees ($^\circ$), where SD is calculated across all tested samples. Results for the ResNet18 are reported from the literature~\cite{Bao2025}. Bold values indicate the best performance for each test dataset.}
    \label{tab:cross_domain}
    \resizebox{5in}{!}{%
    \begin{tabular}{l|ccc|ccc}
    \toprule
    \textbf{Model} & \multicolumn{3}{c|}{\textbf{ResNet18}} & \multicolumn{3}{c}{\textbf{XGBoost}}  \\
    \midrule 
    \diagbox{\textbf{Train}}{\textbf{Test}} & \textbf{GazeGene} & \textbf{ETH-Xgaze} & \textbf{Gaze360} & \textbf{GazeGene} & \textbf{ETH-Xgaze} & \textbf{Gaze360} \\
    \midrule
    \textbf{GazeGene} & - & \textbf{12.87} & 24.20 & - & 17.56 (10.51) & 17.31 (12.52)  \\
    \textbf{ETH-Xgaze} & 15.42 & - & 19.47 & 18.83 (10.99) & - & 24.28 (13.71)  \\
    \textbf{Gaze360} & 18.80 & 18.51 & - & 25.89 (15.95) & 24.88 (17.10) & - \\
    \bottomrule
    \toprule
    \textbf{Model} & \multicolumn{3}{c|}{\textbf{Holistic MLP}} & \multicolumn{3}{c}{\textbf{Siamese MLP}} \\
    \midrule 
    \diagbox{\textbf{Train}}{\textbf{Test}} & \textbf{GazeGene} & \textbf{ETH-Xgaze} & \textbf{Gaze360} & \textbf{GazeGene} & \textbf{ETH-Xgaze} & \textbf{Gaze360} \\
    \midrule
    \textbf{GazeGene} &  - & 15.35 (11.00) & 17.29 (12.95) & - & 13.85 (9.87) & \textbf{17.28 (13.04)} \\
    \textbf{ETH-Xgaze} & 15.16 (10.10) & - & 21.86 (14.78) & \textbf{14.85 (10.03)} & - & 21.27 (14.67) \\
    \textbf{Gaze360} & 17.13 (11.52) & 20.13 (13.03) & - & 15.99 (10.81) & 17.55 (11.38) & - \\
    \bottomrule
    \end{tabular}%
    }
\end{table*}

\subsection{Feature importance analysis}
\label{sec:feature importance}
To interpret the models trained on landmark-based datasets, we analyze the contribution of different facial landmarks to the gaze estimation task using Permutation Feature Importance (PFI). This method measures the dependence of the model on specific features by evaluating the performance degradation when those features are randomly permuted. In the following, we limit our discussion to the siamese MLP, as it achieves the best within- and cross-domain results among the three landmark-based approaches.

\begin{figure}[!ht]
    \centering
    \includegraphics[width=2.5in]{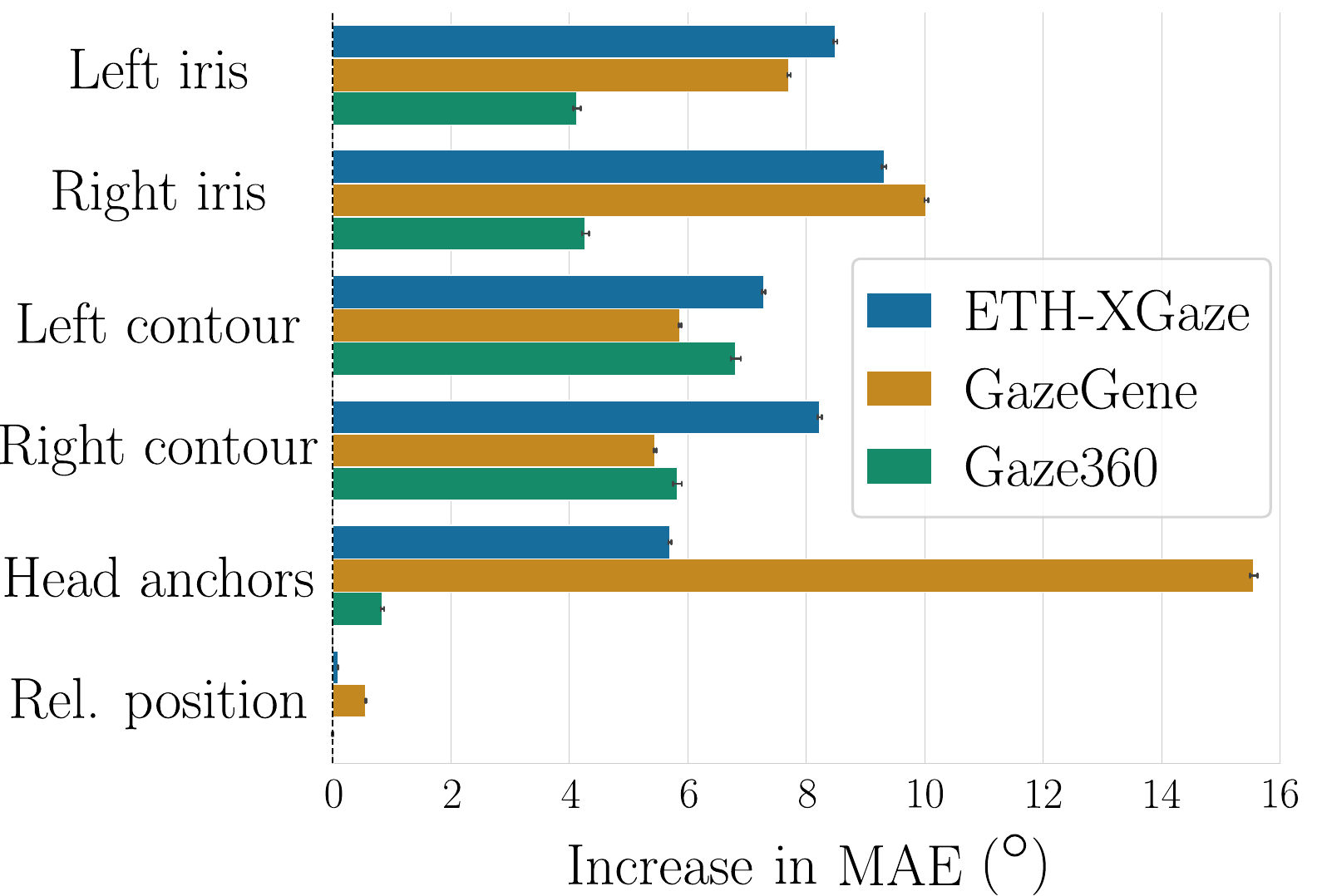}
    \caption{Permutation Feature Importance (PFI) analysis of the siamese MLP trained on ETH-Xgaze, GazeGene, and Gaze360. The importance of each feature group is measured by the increase in mean angular error ($^\circ$) when the corresponding input features are permuted while keeping others constant. Higher values indicate a greater reliance of the model on that specific feature group. Error bars represent the standard deviation over 1000 permutation cycles.}
    \label{fig:feature_importance}
\end{figure}

We define feature groups based on the semantic structure of the input data: the iris (pupil center and iris contour), the eye contour (eyelids and corners), the head anchors (glabella and nose bridge), and the relative position vector. For any feature group $\mathbf{f}$, we construct a validation set $\mathcal{D}_f$ by permuting the values of the features within $\mathbf{f}$ in the original validation set $\mathcal{D}$, while keeping all other features and target labels fixed.
We define the importance score $I_f$ as the increase in the Mean Angular Error (MAE) relative to the baseline performance, i.e.,
\begin{equation}\label{eq:importance score}
    I_f = \bar{E}\paren{\mathcal{D}_f} -\bar{E}\paren{\mathcal{D}}
\end{equation}
where $\bar{E}(\cdot)$ represents the mean of the loss function~\eqref{eq:loss function} on the given dataset. We repeat the permutation procedure 1000 times for each feature group and report the results in Fig.~\ref{fig:feature_importance}. For GazeGene, the head anchors are the most critical feature group, followed by the irises with a bias towards the right one (likely due to the biased dataset distribution, as shown in Fig.~\ref{fig:distributions}). For ETH-XGaze, the irises and eye contours are the most important features, with the head anchors playing a secondary role. Finally, for Gaze360, the eye contours emerge as the primary contributors, followed by the irises.

\section{Discussion}
\label{sec:discussion}
We find that landmark-based models underperform appearance-based approaches in within-domain evaluations. We hypothesize that the lower within-domain performance is due to noise introduced during the dataset extraction, particularly inaccuracies in face detection and head pose estimation, which degrade the quality and coverage of the original training data, as shown in Section~\ref{sec:res_data}. 
However, landmark-based MLP networks achieve cross-domain performances that are comparable to those of the computationally heavier ResNet18 baseline. Therefore, MLPs experience a lower performance drop with respect to ResNet18, thus exhibiting an overall stronger cross-domain generalization. 
This suggests that facial landmarks encode the essential geometric information for accurate gaze estimation, while being less sensitive to domain-specific variations. Indeed, by operating on sparse geometric coordinates rather than on pixel intensities, landmark-based models are inherently invariant to factors such as illumination and skin tone differences.

The siamese MLP, which explicitly models the binocular structure of gaze, produces the best results among landmark-based methods, suggesting that architectures designed to reflect the underlying geometry can better capture generalized gaze rules than holistic or tree-based approaches.

Beyond accuracy and generalization, the landmark-based framework offers practical benefits. First, replacing a deep convolutional backbone with simple MLPs dramatically reduces the computational cost during regression, making it ideal for devices with lower computational capabilities; in fact, both approaches require an initial detection and normalization step, but the one based on landmarks removes the need for a heavy feature extractor after normalization. Second, storing or processing only landmark coordinates might also reduce the exposure of sensitive biometric data.
Third, the landmark-based approach provides greater interpretability of trained models.
In particular, by a Permutation Feature Importance (PFI) analysis, we show that models rely on different features, depending on the training dataset. For example, the model trained on GazeGene shows a bias towards the right eye, likely reflecting the underlying distribution of the synthetic data. This interpretability might serve as a diagnostic tool, allowing researchers to effectively assess the quality and biases of gaze datasets.
The primary bottleneck of the proposed approach is the accuracy and robustness of the upstream landmark detector. 
As observed in the analysis of the extracted datasets, the detector fails or produces unreliable results at extreme gaze and head pose angles, leading to a reduction in the actual quality, range and size of the datasets. For the proposed pipeline, the retention rate is between 73\% and 82\% across the three considered datasets (Gaze360, GazeGene and ETH-XGaze).
Moreover, inaccuracies in landmark detection propagate into the normalization procedure, likely contributing to the performance gap observed between our models and the ResNet18 baseline in within-domain tests. Therefore, we expect that future improvements in the reliability of landmark detectors would directly translate to performance gains in this framework.
In conclusion, our results support the fact that synthetic data can be used successfully for geometric gaze estimators. In fact, in the landmark-based approach, photorealistic images are not necessary, as long as geometric features are preserved. Therefore, future work should focus on enhancing landmark detection and expanding datasets to cover broader geometric and demographic variation without requiring photorealism.

\section{Conclusion}
\label{sec:conclusion}
This study supports the viability of estimating gaze direction solely from facial landmarks. By evaluating lightweight MLPs and decision tree ensembles against state-of-the-art appearance-based baselines, we show that geometric features alone contain sufficient information for accurate gaze estimation. Although current landmark detectors introduce noise that limits within-domain precision, the proposed siamese MLP architecture exhibits comparable generalization performances of significantly heavier ResNet18 models. These findings highlight a clear path toward hardware-efficient and privacy-preserving gaze tracking. By decoupling the regression task from raw pixel data, we reduce the computational cost and exposure of sensitive biometric information.

\section*{Data and code availability}
Source code and extracted landmark datasets are available at \url{https://github.com/daniele-agostinelli/LandmarkGaze.git}.

\bibliographystyle{IEEEtran}
\bibliography{references}

@article{Cheng2024,
  author =        {Cheng, Yihua and Wang, Haofei and Bao, Yiwei and
                   Lu, Feng},
  journal =       {IEEE Transactions on Pattern Analysis and Machine
                   Intelligence},
  number =        {12},
  pages =         {7509-7528},
  title =         {Appearance-Based Gaze Estimation With Deep Learning:
                   A Review and Benchmark},
  volume =        {46},
  year =          {2024},
  doi =           {10.1109/TPAMI.2024.3393571},
}

@article{Guestrin2006,
  author =        {Guestrin, E.D. and Eizenman, M.},
  journal =       {IEEE Transactions on Biomedical Engineering},
  number =        {6},
  pages =         {1124-1133},
  title =         {General theory of remote gaze estimation using the
                   pupil center and corneal reflections},
  volume =        {53},
  year =          {2006},
  doi =           {10.1109/TBME.2005.863952},
}

@article{Hansen2009,
  author =        {Hansen, Dan Witzner and Ji, Qiang},
  journal =       {IEEE Transactions on Pattern Analysis and Machine
                   Intelligence},
  number =        {3},
  pages =         {478-500},
  title =         {In the Eye of the Beholder: A Survey of Models for
                   Eyes and Gaze},
  volume =        {32},
  year =          {2010},
  doi =           {10.1109/TPAMI.2009.30},
}

@inproceedings{Takemura2017,
  author =        {Takemura, Kentaro and Yamagishi, Kenta},
  booktitle =     {2017 IEEE International Conference on Systems, Man,
                   and Cybernetics (SMC)},
  pages =         {1529–1534},
  publisher =     {IEEE Press},
  title =         {A hybrid eye-tracking method using a multispectral
                   camera},
  year =          {2017},
  doi =           {10.1109/SMC.2017.8122831},
  url =           {https://doi.org/10.1109/SMC.2017.8122831},
}

@inproceedings{Zhang2015,
  author =        {Zhang, Xucong and Sugano, Yusuke and Fritz, Mario and
                   Bulling, Andreas},
  booktitle =     {Proceedings of the IEEE Conference on Computer Vision
                   and Pattern Recognition (CVPR)},
  month =         {June},
  title =         {Appearance-Based Gaze Estimation in the Wild},
  year =          {2015},
}

@article{Cheng2020,
  author =        {Cheng, Yihua and Zhang, Xucong and Lu, Feng and
                   Sato, Yoichi},
  journal =       {IEEE Transactions on Image Processing},
  number =        {},
  pages =         {5259-5272},
  title =         {Gaze Estimation by Exploring Two-Eye Asymmetry},
  volume =        {29},
  year =          {2020},
  doi =           {10.1109/TIP.2020.2982828},
}

@inproceedings{Park2018b,
  author =        {Park, Seonwook and Spurr, Adrian and Hilliges, Otmar},
  booktitle =     {Proceedings of the European Conference on Computer
                   Vision (ECCV)},
  month =         {September},
  title =         {Deep Pictorial Gaze Estimation},
  year =          {2018},
}

@inproceedings{Park2019,
  author =        {Park, Seonwook and Mello, Shalini De and
                   Molchanov, Pavlo and Iqbal, Umar and Hilliges, Otmar and
                   Kautz, Jan},
  booktitle =     {Proceedings of the IEEE/CVF International Conference
                   on Computer Vision (ICCV)},
  month =         {October},
  title =         {Few-Shot Adaptive Gaze Estimation},
  year =          {2019},
}

@inproceedings{Zhang2017,
  author =        {Zhang, Xucong and Sugano, Yusuke and Fritz, Mario and
                   Bulling, Andreas},
  booktitle =     {Proceedings of the IEEE Conference on Computer Vision
                   and Pattern Recognition (CVPR) Workshops},
  month =         {July},
  title =         {It's Written All Over Your Face: Full-Face
                   Appearance-Based Gaze Estimation},
  year =          {2017},
}

@inproceedings{Chen2019,
  address =       {Cham},
  author =        {Chen, Zhaokang and Shi, Bertram E.},
  booktitle =     {Computer Vision -- ACCV 2018},
  editor =        {Jawahar, C.V. and Li, Hongdong and Mori, Greg and
                   Schindler, Konrad},
  pages =         {309--324},
  publisher =     {Springer International Publishing},
  title =         {Appearance-Based Gaze Estimation Using
                   Dilated-Convolutions},
  year =          {2019},
  isbn =          {978-3-030-20876-9},
}

@article{Cheng2020b,
  author =        {Cheng, Yihua and Huang, Shiyao and Wang, Fei and
                   Qian, Chen and Lu, Feng},
  journal =       {Proceedings of the AAAI Conference on Artificial
                   Intelligence},
  month =         {Apr.},
  number =        {07},
  pages =         {10623-10630},
  title =         {A Coarse-to-Fine Adaptive Network for
                   Appearance-Based Gaze Estimation},
  volume =        {34},
  year =          {2020},
  doi =           {10.1609/aaai.v34i07.6636},
  url =           {https://ojs.aaai.org/index.php/AAAI/article/view/6636},
}

@inproceedings{Bao2021,
  author =        {Bao, Yiwei and Cheng, Yihua and Liu, Yunfei and
                   Lu, Feng},
  booktitle =     {2020 25th International Conference on Pattern
                   Recognition (ICPR)},
  number =        {},
  pages =         {9936-9943},
  title =         {Adaptive Feature Fusion Network for Gaze Tracking in
                   Mobile Tablets},
  volume =        {},
  year =          {2021},
  doi =           {10.1109/ICPR48806.2021.9412205},
}

@inproceedings{Cheng2022,
  author =        {Cheng, Yihua and Lu, Feng},
  booktitle =     {2022 26th International Conference on Pattern
                   Recognition (ICPR)},
  number =        {},
  pages =         {3341-3347},
  title =         {Gaze Estimation using Transformer},
  volume =        {},
  year =          {2022},
  doi =           {10.1109/ICPR56361.2022.9956687},
}

@inproceedings{Liu2021,
  author =        {Liu, Yunfei and Liu, Ruicong and Wang, Haofei and
                   Lu, Feng},
  booktitle =     {Proceedings of the IEEE/CVF International Conference
                   on Computer Vision (ICCV)},
  month =         {October},
  pages =         {3835-3844},
  title =         {Generalizing Gaze Estimation With Outlier-Guided
                   Collaborative Adaptation},
  year =          {2021},
}

@inproceedings{Wang2022,
  author =        {Wang, Yaoming and Jiang, Yangzhou and Li, Jin and
                   Ni, Bingbing and Dai, Wenrui and Li, Chenglin and
                   Xiong, Hongkai and Li, Teng},
  booktitle =     {Proceedings of the IEEE/CVF Conference on Computer
                   Vision and Pattern Recognition (CVPR)},
  month =         {June},
  pages =         {19376-19385},
  title =         {Contrastive Regression for Domain Adaptation on Gaze
                   Estimation},
  year =          {2022},
}

@inproceedings{Bao2022,
  author =        {Bao, Yiwei and Liu, Yunfei and Wang, Haofei and
                   Lu, Feng},
  booktitle =     {Proceedings of the IEEE/CVF Conference on Computer
                   Vision and Pattern Recognition (CVPR)},
  month =         {June},
  pages =         {4207-4216},
  title =         {Generalizing Gaze Estimation With Rotation
                   Consistency},
  year =          {2022},
}

@inproceedings{Cai2023,
  author =        {Cai, Xin and Zeng, Jiabei and Shan, Shiguang and
                   Chen, Xilin},
  booktitle =     {Proceedings of the IEEE/CVF Conference on Computer
                   Vision and Pattern Recognition (CVPR)},
  month =         {June},
  pages =         {22035-22045},
  title =         {Source-Free Adaptive Gaze Estimation by Uncertainty
                   Reduction},
  year =          {2023},
}

@inproceedings{Bao2024,
  author =        {Bao, Yiwei and Lu, Feng},
  booktitle =     {Proceedings of the IEEE/CVF Conference on Computer
                   Vision and Pattern Recognition (CVPR)},
  month =         {June},
  pages =         {1409-1418},
  title =         {From Feature to Gaze: A Generalizable Replacement of
                   Linear Layer for Gaze Estimation},
  year =          {2024},
}

@article{Cheng2022b,
  author =        {Cheng, Yihua and Bao, Yiwei and Lu, Feng},
  journal =       {Proceedings of the AAAI Conference on Artificial
                   Intelligence},
  month =         {Jun.},
  number =        {1},
  pages =         {436-443},
  title =         {PureGaze: Purifying Gaze Feature for Generalizable
                   Gaze Estimation},
  volume =        {36},
  year =          {2022},
  doi =           {10.1609/aaai.v36i1.19921},
  url =           {https://ojs.aaai.org/index.php/AAAI/article/view/19921},
}

@article{Xu2023,
  author =        {Xu, Mingjie and Wang, Haofei and Lu, Feng},
  journal =       {Proceedings of the AAAI Conference on Artificial
                   Intelligence},
  month =         {Jun.},
  number =        {3},
  pages =         {3027-3035},
  title =         {Learning a Generalized Gaze Estimator from
                   Gaze-Consistent Feature},
  volume =        {37},
  year =          {2023},
  doi =           {10.1609/aaai.v37i3.25406},
  url =           {https://ojs.aaai.org/index.php/AAAI/article/view/25406},
}

@inproceedings{Liang2024,
  address =       {Cham},
  author =        {Liang, Ziyang and Bao, Yiwei and Lu, Feng},
  booktitle =     {Computer Vision -- ECCV 2024},
  editor =        {Leonardis, Ale{\v{s}} and Ricci, Elisa and
                   Roth, Stefan and Russakovsky, Olga and
                   Sattler, Torsten and Varol, G{\"u}l},
  pages =         {219--235},
  publisher =     {Springer Nature Switzerland},
  title =         {De-confounded Gaze Estimation},
  year =          {2025},
  isbn =          {978-3-031-73337-6},
}

@inproceedings{Zhang2018,
  address =       {New York, NY, USA},
  author =        {Zhang, Xucong and Sugano, Yusuke and
                   Bulling, Andreas},
  booktitle =     {Proceedings of the 2018 ACM Symposium on Eye Tracking
                   Research \& Applications},
  publisher =     {Association for Computing Machinery},
  series =        {ETRA '18},
  title =         {Revisiting data normalization for appearance-based
                   gaze estimation},
  year =          {2018},
  doi =           {10.1145/3204493.3204548},
  isbn =          {9781450357067},
  url =           {https://doi.org/10.1145/3204493.3204548},
}

@inproceedings{Krafka2016,
  author =        {Krafka, Kyle and Khosla, Aditya and Kellnhofer, Petr and
                   Kannan, Harini and Bhandarkar, Suchendra and
                   Matusik, Wojciech and Torralba, Antonio},
  booktitle =     {Proceedings of the IEEE Conference on Computer Vision
                   and Pattern Recognition (CVPR)},
  month =         {June},
  title =         {Eye Tracking for Everyone},
  year =          {2016},
}

@inproceedings{Yu2018,
  author =        {Yu, Yu and Liu, Gang and Odobez, Jean-Marc},
  booktitle =     {Proceedings of the European Conference on Computer
                   Vision (ECCV) Workshops},
  month =         {September},
  title =         {Deep Multitask Gaze Estimation with a Constrained
                   Landmark-Gaze Model},
  year =          {2018},
}

@article{Lei2023,
  address =       {New York, NY, USA},
  author =        {Lei, Yaxiong and He, Shijing and Khamis, Mohamed and
                   Ye, Juan},
  journal =       {ACM Comput. Surv.},
  month =         sep,
  number =        {2},
  publisher =     {Association for Computing Machinery},
  title =         {An End-to-End Review of Gaze Estimation and its
                   Interactive Applications on Handheld Mobile Devices},
  volume =        {56},
  year =          {2023},
  doi =           {10.1145/3606947},
  issn =          {0360-0300},
  url =           {https://doi.org/10.1145/3606947},
}

@article{agostinelli2026,
  author =        {Agostinelli, Thomas and Generosi, Andrea and
                   Mengoni, Maura},
  journal =       {The International Journal of Advanced Manufacturing
                   Technology},
  pages =         {1--24},
  publisher =     {Springer},
  title =         {A novel approach for monocular RGB-based ergonomics
                   monitoring in industrial workspaces employing
                   synthetic datasets to train a deep learning model},
  year =          {2026},
}

@inproceedings{macedo2024,
  author =        {Macedo, Alessandra Alaniz and Persona, Leandro and
                   Meloni, Fernando},
  booktitle =     {Brazilian Symposium on Multimedia and the Web
                   (WebMedia)},
  organization =  {SBC},
  pages =         {257--266},
  title =         {Recognition of emotions through facial geometry with
                   normalized landmarks},
  year =          {2024},
}

@article{kumar2025,
  author =        {Kumar, Akhilesh and Kumar, Awadhesh and Gupta, Sumit},
  journal =       {SN Computer Science},
  number =        {2},
  pages =         {120},
  publisher =     {Springer},
  title =         {Machine learning-driven emotion recognition through
                   facial landmark analysis},
  volume =        {6},
  year =          {2025},
}

@inproceedings{Chen2008,
  author =        {Chen, Jixu and Ji, Qiang},
  booktitle =     {2008 19th International Conference on Pattern
                   Recognition},
  number =        {},
  pages =         {1-4},
  title =         {3D gaze estimation with a single camera without IR
                   illumination},
  volume =        {},
  year =          {2008},
  doi =           {10.1109/ICPR.2008.4761343},
}

@inproceedings{Wang2017,
  author =        {Wang, Kang and Ji, Qiang},
  booktitle =     {Proceedings of the IEEE International Conference on
                   Computer Vision (ICCV)},
  month =         {Oct},
  title =         {Real Time Eye Gaze Tracking With 3D Deformable
                   Eye-Face Model},
  year =          {2017},
}

@inproceedings{Park2018,
  address =       {New York, NY, USA},
  author =        {Park, Seonwook and Zhang, Xucong and Bulling, Andreas and
                   Hilliges, Otmar},
  booktitle =     {Proceedings of the 2018 ACM Symposium on Eye Tracking
                   Research \& Applications},
  publisher =     {Association for Computing Machinery},
  series =        {ETRA '18},
  title =         {Learning to find eye region landmarks for remote gaze
                   estimation in unconstrained settings},
  year =          {2018},
  doi =           {10.1145/3204493.3204545},
  isbn =          {9781450357067},
  url =           {https://doi.org/10.1145/3204493.3204545},
}

@article{Oh2022,
  author =        {Oh, Jaekwang and Lee, Youngkeun and Yoo, Jisang and
                   Kwon, Soonchul},
  journal =       {Sensors (Basel)},
  month =         may,
  number =        {11},
  pages =         {4026},
  publisher =     {MDPI AG},
  title =         {Improved feature-based gaze estimation using
                   self-attention module and synthetic eye images},
  volume =        {22},
  year =          {2022},
  language =      {en},
}

@article{Her2023,
  author =        {Her, Paris and Manderle, Logan and Dias, Philipe A. and
                   Medeiros, Henry and Odone, Francesca},
  journal =       {IEEE Transactions on Image Processing},
  number =        {},
  pages =         {2335-2347},
  title =         {Uncertainty-Aware Gaze Tracking for Assisted Living
                   Environments},
  volume =        {32},
  year =          {2023},
  doi =           {10.1109/TIP.2023.3253253},
}

@misc{ye2023,
  author =        {Esther Enhui Ye and John Enzhou Ye and Joseph Ye and
                   Jacob Ye and Runzhou Ye},
  title =         {Low-cost Geometry-based Eye Gaze Detection using
                   Facial Landmarks Generated through Deep Learning},
  year =          {2023},
  url =           {https://arxiv.org/abs/2401.00406},
}

@inproceedings{Wood2016,
  address =       {New York, NY, USA},
  author =        {Wood, Erroll and Baltru\v{s}aitis, Tadas and
                   Morency, Louis-Philippe and Robinson, Peter and
                   Bulling, Andreas},
  booktitle =     {Proceedings of the Ninth Biennial ACM Symposium on
                   Eye Tracking Research \& Applications},
  pages =         {131–138},
  publisher =     {Association for Computing Machinery},
  series =        {ETRA '16},
  title =         {Learning an appearance-based gaze estimator from one
                   million synthesised images},
  year =          {2016},
  doi =           {10.1145/2857491.2857492},
  isbn =          {9781450341257},
  url =           {https://doi.org/10.1145/2857491.2857492},
}

@inproceedings{Kellnhofer2019,
  author =        {Kellnhofer, Petr and Recasens, Adria and Stent, Simon and
                   Matusik, Wojciech and Torralba, Antonio},
  booktitle =     {Proceedings of the IEEE/CVF International Conference
                   on Computer Vision (ICCV)},
  month =         {October},
  title =         {Gaze360: Physically Unconstrained Gaze Estimation in
                   the Wild},
  year =          {2019},
}

@inproceedings{Zhang2020,
  address =       {Cham},
  author =        {Zhang, Xucong and Park, Seonwook and Beeler, Thabo and
                   Bradley, Derek and Tang, Siyu and Hilliges, Otmar},
  booktitle =     {Computer Vision -- ECCV 2020},
  editor =        {Vedaldi, Andrea and Bischof, Horst and Brox, Thomas and
                   Frahm, Jan-Michael},
  pages =         {365--381},
  publisher =     {Springer International Publishing},
  title =         {ETH-XGaze: A Large Scale Dataset for Gaze Estimation
                   Under Extreme Head Pose and Gaze Variation},
  year =          {2020},
  isbn =          {978-3-030-58558-7},
}

@inproceedings{Bao2025,
  author =        {Bao, Yiwei and Wang, Zhiming and Lu, Feng},
  booktitle =     {Proceedings of the IEEE/CVF Conference on Computer
                   Vision and Pattern Recognition},
  title =         {GazeGene: Large-scale Synthetic Gaze Dataset with 3D
                   Eyeball Annotations},
  year =          {2025},
}

@inproceedings{Lugaresi2019,
  author =        {Camillo Lugaresi and Jiuqiang Tang and Hadon Nash and
                   Chris McClanahan and Esha Uboweja and Michael Hays and
                   Fan Zhang and Chuo-Ling Chang and Ming Yong and
                   Juhyun Lee and Wan-Teh Chang and Wei Hua and
                   Manfred Georg and Matthias Grundmann},
  booktitle =     {Third Workshop on Computer Vision for AR/VR at IEEE
                   Computer Vision and Pattern Recognition (CVPR) 2019},
  title =         {MediaPipe: A Framework for Perceiving and Processing
                   Reality},
  year =          {2019},
  url =           {https://mixedreality.cs.cornell.edu/s/
                  NewTitle_May1_MediaPipe_CVPR_CV4ARVR_Workshop_2019.pdf},
}

@article{Marchand2016,
  author =        {Marchand, Eric and Uchiyama, Hideaki and
                   Spindler, Fabien},
  journal =       {{IEEE Transactions on Visualization and Computer
                   Graphics}},
  month =         dec,
  number =        {12},
  pages =         {2633 - 2651},
  publisher =     {{Institute of Electrical and Electronics Engineers}},
  title =         {{Pose Estimation for Augmented Reality: A Hands-On
                   Survey}},
  volume =        {22},
  year =          {2016},
  doi =           {10.1109/TVCG.2015.2513408},
  url =           {https://inria.hal.science/hal-01246370},
}

@article{opencv_library,
  author =        {Bradski, G.},
  journal =       {Dr. Dobb's Journal of Software Tools},
  title =         {{The OpenCV Library}},
  year =          {2000},
}

\end{document}